\documentclass{ecai}
\usepackage{graphicx}
\usepackage{latexsym}
\usepackage{amsfonts}
\usepackage{amsmath}
\usepackage{algpseudocode}
\usepackage{algorithm}
\usepackage{multirow}
\usepackage{float}
\usepackage{graphicx}
\usepackage{setspace} 
\usepackage{url}
\usepackage{latexsym}
\usepackage{amsfonts}
\usepackage{amsmath}
\usepackage{algpseudocode}
\usepackage{algorithm}
\usepackage{multirow}
\usepackage{float}
\usepackage{graphicx}
\usepackage{xcolor}
\usepackage[capitalize,noabbrev]{cleveref}
\usepackage[compact]{titlesec}         
\titlespacing{\section}{8pt}{8pt}{8pt} 
\begin{document}

\begin{frontmatter}
\title{\fontsize{24pt}{28pt}\selectfont Using the Nucleolus for Incentive Allocation in Vertical Federated Learning}


\author{\fnms{Afsana}~\snm{Khan}}
\author{\fnms{Marijn}~\snm{ten Thij}}
\author{\fnms{Frank}~\snm{Thuijsman}}
\author{\fnms{Anna}~\snm{Wilbik}}

\address{Department of Advanced Computing Sciences\\Maastricht University\\Netherlands\\\textit{Corresponding Author: a.khan@maastrichtuniversity.nl}}
 

\begin{abstract}
Vertical federated learning (VFL) is a promising approach for collaboratively training machine learning models using private data partitioned vertically across different parties. Ideally in a VFL setting, the active party (party possessing features of samples with labels) benefits by improving its machine learning model through collaboration with some passive parties (parties possessing additional features of the same samples without labels) in a privacy-preserving manner. However, motivating passive parties to participate in VFL can be challenging. In this paper, we focus on the problem of allocating incentives to the passive parties by the active party based on their contributions to the VFL process. We address this by formulating the incentive allocation problem as a bankruptcy game, a concept from cooperative game theory. Using the Talmudic division rule, which leads to the \textit{Nucleolus} as its solution, we ensure a fair distribution of incentives. We evaluate our proposed method on synthetic and real-world datasets and show that it ensures fairness and stability in incentive allocation among passive parties who contribute their data to the federated model. Additionally, we compare our method to the existing solution of calculating Shapley values and show that our approach provides a more efficient solution with fewer computations.
\end{abstract}
\end{frontmatter}

\section{Introduction}
Federated Learning (FL) \cite{bonawitz2019towards}, a form of distributed learning enables privacy-preserving model training across decentralized data, avoiding the need to pool data centrally.  FL enhances model performance by leveraging diverse data across different entities. It is categorized into Horizontal FL (HFL) for shared features across distinct samples, Vertical FL (VFL) for varied features within identical samples, and Hybrid FL combining both. Each of these types has its unique application areas based on the nature of the data and the specific requirements of the task \cite{yang2019federated}.

Motivating data owners to participate in federated learning is essential for its success. Incentives encourage data owners to contribute data, improving model performance with diverse and representative data sources. Shi et al. \cite{shi2022fedfaim} proposed a framework in an HFL setting, using Shapley values to allocate incentives based on contributions, with a reputation-driven reward allocation policy. Similarly, Zhang et al. \cite{zhang2021incentive} proposed a reward system in HFL where clients bid prices and a selector chooses participants based on contribution quality, with reputation and payment dependent on the quality of contributions. Incentive mechanisms differ between HFL and VFL due to the different roles of data owners and model goals. Unlike an HFL setting, there are two types of data owners or parties involved in the federation in VFL: (1) \textbf{Passive/Host Party}, owning data (features) without labels for model training; and (2) \textbf{Active/Guest Party}, holding labeled data for training models using host data. Hence, the active party in VFL substantially benefits from additional data provided by the passive parties, necessitating a motivating mechanism for the latter to contribute data. However, most of the existing approaches for incentive allocation in federated learning deal with horizontally partitioned data, where the data is partitioned based on the samples. In contrast, vertically partitioned data is partitioned based on the features or attributes. To the best of our knowledge, no existing studies have formulated the incentive allocation problem in VFL as a bankruptcy game, which offers a unique perspective for addressing this challenge in certain scenarios. Our main contributions in this paper include:
\begin{itemize}
    \item Proposing a novel incentive allocation method in VFL by formulating the problem as a bankruptcy game and employing \emph{nucleolus} for allocation, thus introducing a fresh perspective on incentive allocation beyond the traditional Shapley value.
    \item Empirical evaluation of the method on synthetic and public datasets, providing evidence of its effectiveness and efficiency.
    \item Assessment of fairness in the incentive allocation process, ensuring equitable outcomes for participants.
\end{itemize}

\section{Background}
In this section, we discuss two key solution concepts in cooperative game theory: Nucleolus and Shapley value, in order to establish the context for discussing related works and our proposed method in this paper.
\\\\
\noindent A cooperative game is defined by a pair $(N,v)$, where $N$ denotes a finite set of players, and $v$ is a characteristic function mapping from the set of all coalitions $2^N$ to the real numbers $\mathbb{R}$, assigning a value to each coalition $S \subseteq N$. This value, $v(S)$, represents the total payoff or utility that the members of the coalition $S$ can achieve by cooperating. Solution concepts like Nucleolus and Shapley value aim to determine how the payoff for the grand coalition, $v(N)$, where all players cooperate, can be fairly distributed among the players to reflect their contributions towards achieving this collective outcome.
\\\\
\noindent \textbf{Nucleolus: }Introduced by Schmeidler in 1969, the \emph{nucleolus} seeks to minimize the dissatisfaction or "unhappiness" among coalitions in cooperative games. It does this by aiming to reduce the excess, which is the difference between the value a coalition can achieve by itself and the sum of payoffs allocated to its members. The excess for a coalition $S$ under a payoff vector $\mathbf{x}$ is defined as:
\begin{equation}
e(S, \mathbf{x}) = v(S) - \sum_{i \in S} x_i
\label{nucleolus_eqn}
\end{equation}
where $x_i$ denotes the individual payoff to player $i$ within that coalition. The goal of the \emph{nucleolus} is to find a payoff distribution that minimizes the maximum discontent among all possible coalitions. This leads to a definition of the \emph{nucleolus} $\mathbf{x^*}$ as the payoff vector that achieves the following:
\begin{equation}
\mathbf{x^*} = \arg\min_{\mathbf{x}} \max_{S \subseteq N} e(S, \mathbf{x})
\end{equation}
\textbf{Shapley value: }The Shapley value allocates payoffs based on a player's average marginal contribution- the additional value a player adds to a coalition beyond the coalition's value without that player. Mathematically, the Shapley value for a player $i$ in a game with a set of players $N$ and a characteristic function $v$ is defined as:
\begin{equation}
\phi_i(N,v) = \sum_{S \subseteq N \setminus \{i\}} \frac{|S|! \cdot (|N| - |S| - 1)!}{|N|!} \left(v(S \cup \{i\}) - v(S)\right),
\label{shapley_eqn}
\end{equation}
where $v(S \cup \{i\}) - v(S)$ represents the marginal contribution of player $i$ to the coalition $S$. 
\\
\noindent \textbf{Related Works:} Liu et al. \cite{liu2022gtg} formulated a federated learning scenario as a cooperative game by modeling the data owners as players who cooperate to train a high-quality machine learning model. In the context of federated learning, considering a game where the players are represented by owners of labeled datasets, it can be formalized as \cite{rozemberczki2022Shapley}:
\[ N = \{(X_i, Y_i) \mid 1 \leq i \leq n\} \]
Here, \(X_i\) and \(Y_i\) are the feature and label sets, respectively, belonging to the \(i^{th}\) data party. Given a specific labeled test set, denoted as \( (X, Y) \), and a coalition \( S \) comprised of a subset of the data belonging to the parties or clients, the machine learning model trained on \( S \) is represented by \( f_S(\cdot) \). The model predicts labels \( Y^S \) when applied to \( X \). The associated payoff for the coalition \( S \) in this game is described by:
\[ v(S) = g(Y, Y^S) \] 
Here, $g(.)$ is the goodness of fit metric. Liu et al. \cite{liu2022gtg} furthermore proposed a Guided Truncation Gradient Shapley (GTG-Shapley) approach to equitably evaluate participants' contributions to federated learning model performance without revealing their private data. The methodology reconstructs FL models from gradient updates for Shapley value computation and employs a guided Monte Carlo sampling strategy allied with truncation to minimize computational expenses. Another incentive mechanism based on the enhanced Shapley value method was presented in \cite{yang2022federated}, which determines income distribution by accommodating multiple influencing factors as weights and using the analytic hierarchy process (AHP) \cite{saaty1988analytic} to derive the corresponding weight value for each factor. Song et al. \cite{song2019profit} introduced a Shapley value-based contribution index for equitable profit distribution among data providers in federated learning, efficiently approximating contributions to the joint model and significantly reducing computation costs. 

The mentioned works including \cite{wang2020principled, wei2020efficient, le2021incentive, zhan2020learning} are applicable to federated learning for horizontally paritioned scenarios and not VFL due to its different nature of data partitioning. While a significant number of incentive mechanisms concentrate on horizontally federated learning (HFL) settings, some initiatives for vertical federated learning (VFL) have also been introduced. For instance, Fan et al. \cite{fan2022fair} introduced a contribution valuation metric \emph{VerFedSV}, while Wang et al. \cite{wang2019measure} applied the Shapley value to VFL to compute the grouped feature importance of parties, both instances illustrating the versatility and applicability of Shapley value in VFL scenarios. Moreover,  DIG-FL, an efficient method for calculating Shapley values for participants in both VFL and HFL has been proposed in \cite{wang2022efficient}, enabling contribution-based participant weighting without model retraining. These studies highlight various approximation techniques aimed at reducing the computational demand of Shapley values. However, such approximations may not always align with the fairness principles established by cooperative game theory, underscoring a trade-off between computational efficiency and adherence to fairness. Lu et al. \cite{lu2022truthful} introduced an incentive mechanism for vertical federated learning based on the assumption that all parties have label data. This approach doesn't fit the usual setup where only the active party has label data, and passive parties contribute features. Therefore, their method may not be suitable for typical scenarios where the goal is to compensate passive parties for their data contributions. Tan et al. \cite{tan2023fraim} introduce FRAIM, targeting incentive mechanisms in VFL by leveraging feature importance and synthetic data to estimate contributions through a reverse auction. However, our method, framed as a bankruptcy game, offers a straightforward and scalable solution, eliminating the need for synthetic data and simplifying incentive allocation, making it especially suitable for scenarios with constrained federation budgets, emphasizing efficiency and simplicity in resource utilization.

\section{Method}
In this section, we propose our approach to incentive allocation in VFL formulating the problem as a bankruptcy game. The choice of a bankruptcy game is predicated on its suitability for scenarios where resources—in this case, incentives—must be distributed among parties who have claims (i.e., contributions to the learning process) that exceed the available resources. This framework is particularly applicable for VFL settings, where the active party, benefiting directly from the collaborative learning process, must determine how to fairly distribute a finite pool of incentives among multiple passive parties, each contributing additional data features that improve the learning model. 
For completeness, we present here the definitions of the bankruptcy game.
\subsection{Reminder On Bankruptcy Game}
A bankruptcy game, based on the bankruptcy problem, arises when an entity's assets (denoted by \(E\)) are insufficient to meet all creditors' claims, \(c_i\), for \(i=1,...,n\). The game is defined by a set of players (claimants) \(C = \{1, 2, \ldots, n\}\), each with a claim \(c_i\) on the estate, where typically \(\sum_{i=1}^{n} c_i > E\). In this context, the coalition value for a subset of claimants \(S \subseteq C\) is determined by computing the remaining estate after fulfilling the claims of claimants not involved in that coalition. 

\begin{equation}
v(S) = \max\{0, E - \sum_{i \in S} c_i\}
\label{bankruptcy_eqn}
\end{equation}

\noindent To solve the allocation problem in bankruptcy games, mostly cooperative game theory solutions such as the nucleolus and the Shapley value are utilized. These solutions require computation of the worth for all possible claimant coalitions, which is derived from the characteristic function of the game [Equation~\eqref{nucleolus_eqn} and~\eqref{shapley_eqn}]. However, solving a bankruptcy game conventionally by the nucleolus or Shapley value leads to the computation of all possible coalition values and is computationally expensive; for \(n\) participants, we would need to compute \(2^n\) coalitions. Nevertheless, in \cite{aumann1985game}, Aumann and Maschler prove that the solution of a bankruptcy problem using the Talmud division rule, also known as Contest-Garment rule \cite{de2008talmud} is the \emph{nucleolus} of its corresponding bankruptcy game. This means that by applying the Talmud division rule to the original bankruptcy problem, one inherently obtains what is mathematically formalized as the nucleolus in the game-theoretic setting, without explicitly constructing the entire game that involves computing values of all possible coalitions. Hence, we employ the nucleolus for solving the allocation problem as it is easier to compute without the need to calculate the worth of every possible coalition, unlike the Shapley value. A bankruptcy problem distributes an estate \(E\) among creditors with claims \(c_1, c_2, \ldots, c_n\) (where \(\sum_{i=1}^{n} c_i \geq E\)), formally defined as the pair \((E;c) \in \mathbb{R} \times \mathbb{R}^{n}\). In addressing the allocation in a bankruptcy problem a division rule \(f\) \((E;c)\) prescribes a solution or payoff \(f(E; c) = (f_{1}(E;c), f_{2}(E;c), \ldots , f_{n}(E;c))\) to each creditor, such that:
\begin{itemize}
    \item \(f_{i}(E;c) \geq 0\) for all \(i \in \{1,2,\ldots,n\}\)
    \item \(\sum_{i=1}^{n} f_{i}(E;c) = E\)
\end{itemize}
\noindent \textbf{Rationale for Talmud Division Rule: }For $n$-person bankruptcy problem, the Talmud division rule is the unique rule for dividing the estate $E$ in a pairwise-consistent way with the following solution for a two-person bankruptcy problem:  $f_i = c_i - 0.5(c_1+c_2 - E)$ for $i=1,2$, i.e.\ they split the joint loss equally. For the $n$-person problem this means that for any pair $(i,j)\in N^2$ the payoffs $(f_i,f_j)$ in the $n$-person problem are the exact equal-split-of-losses solution of the 2-person problem with $E=f_i + f_j$ and claims $c_i, c_j$. \\ 
This rule is particularly significant because it minimizes the maximum excess (gain/loss) across all parties and it is the unique n-party solution that is pairwise consistent with sharing losses equally in a two-party problem. According to this rule, small creditors always lose less than larger creditors, and small creditors always get less than larger creditors. \cref{alg:talmud} explains how the Talmud division rule is applied in the bankruptcy problem.
\begin{algorithm}
\caption{Talmud Division Rule for Bankruptcy Problem}
\label{alg:talmud}
\begin{algorithmic}
\setstretch{0.8} 
\State Order the creditors from lowest to highest claim.
\State Allocate equally among all creditors until the lowest creditor receives half of their claim.
\While{remaining estate is not empty and not all creditors have received half of their claim}
    \State Remove the lowest creditor.
    \State Allocate the remaining estate equally among remaining creditors until the new lowest creditor receives half of their claim.
\EndWhile
\State Order all creditors from highest to lowest claim.
\While{remaining estate is not empty}
    \State Allocate the remaining estate equally to highest creditors until their loss (claim minus amount received) equals the loss of the next creditor.
    \State Include the next creditor in the group of highest creditors.
\EndWhile
\end{algorithmic}
\end{algorithm}

\subsection{Problem Formulation}
In a VFL setting the active party improves its model by using data from passive parties. We view this scenario as a bankruptcy problem where the passive parties have stakes in the performance gains of the federated model. The ``estate" in this case is the improvement in performance over the active party's individual contribution. We measure a passive party's claim as the boost in performance from their unique data contribution.\\
In a VFL setup with \(N\) parties, let us consider a scenario where a federated machine learning model, \(M_f\), is being trained. The objective here is to minimize a function, \(F(M_f)\), over the data of all participating parties. The model \(M_f\) represents the federated model that is obtained through the collective contributions of all parties involved in the setup. Let \(M_a\) denote the local model of the active party \(a\). This model is trained exclusively on its local data, \(D_a\), without any external collaboration. We evaluate the performance of \(M_a\) using a designated evaluation metric, \(R\). Similarly, the performance of the federated model, \(M_f\), is also assessed using the metric \(R\). Consequently, we define the term 'estate', represented by \(E\), as the difference in performance between the federated model \(M_f\) and the local model \(M_a\):

\begin{equation}
\label{estate_eqn}
E = R(M_f) - R(M_a)
\end{equation}

\noindent For each passive party \(i\) that wishes to claim credit for its contribution to \(M_f\), we calculate its claim, \(C_i\). This is done by comparing the performance of \(M_f\) when party \(i\) collaborates solely with the active party \(a\). The model resulting from this exclusive collaboration is denoted as \(M_{ai}\), and its performance is also measured using \(R\). Thus, the claim of party \(i\), \(C_i\), is computed as:

\begin{equation}
\label{claims_eqn}
C_i = R(M_{ai}) - R(M_a) 
\end{equation}

\noindent Here, \(M_{ai}\) is trained using the data from both parties \(a\) and \(i\), while \(M_a\) is trained only on the data from party \(a\).

\subsection{Incentive Allocation}

Upon calculating the claims from each passive party and determining the overall estate of the VFL, our proposed method solves the bankruptcy problem using Talmud's division rule to allocate incentives fairly among the passive parties.

\noindent In this context, the incentive allocation problem in VFL can be viewed as a bankruptcy problem where the claims are represented by a vector \(C = [C_1, C_2, ..., C_N]\), and \(E\) is the total estate available for distribution. \\ The payoff vector is represented as:

\begin{equation}
P = [p_1, p_2, ..., p_N] \quad \text{such that} \quad \sum_{i=1}^{N} p_{i} = E
\label{eqn:6}
\end{equation}

\noindent This vector \(P\) signifies the incentive allocated to each passive party, which corresponds to the \emph{nucleolus} of the bankruptcy game. Moreover, in scenarios where the total estate \(E\) is equal to or greater than the sum of all claims (\(E \geq \sum_{i=1}^{N} C_i\)), each passive party is allocated their full claim (\(p_i = C_i\)), reflecting a non-bankruptcy scenario where the available resources suffice to satisfy all parties' claims in full. This ensures that the incentive allocation remains fair and equitable, consistent with the principles underlying the Talmud's division rule, even in situations where the estate is not limited.

\section{Experimental Setup}
This section describes the experimental setup designed to evaluate the proposed method. For simplicity, a binary classification problem was chosen. The proposed method was evaluated on a synthetic and three public datasets from the UCI repository with varying sample sizes and feature dimensions.
\subsection{Datasets}
The datasets were pre-processed to address issues such as missing values and duplicates. Additionally, categorical features in the datasets were encoded using One-Hot-Encoding. 
\begin{itemize}
    \item \textbf{Synthetic Dataset (10,000 instances, 20 features):} Generated using scikit-learn's \texttt{make\_classification} for binary classification, with added Gaussian noise to mimic real-world data complexity.
    \item \textbf{Heart Disease Dataset (303 instances, 13 features):} This smaller dataset provides a challenging test case for predicting the presence of heart disease, focusing on medical features.
    \item \textbf{Bank Marketing Dataset (11,162 instances, 15 features):} Representing a medium-sized dataset, it offers insights into a Portuguese bank's marketing campaigns, predicting client subscription to term deposits.
    \item \textbf{Spam Emails Dataset (4210 instances, 57 features):} With a larger feature set, this dataset is used for spam email classification based on word frequency, char frequency, and header features.
\end{itemize}
\subsection{VFL Training}
To simulate a vertically federated environment, the datasets were partitioned vertically, with each partition representing the local data held by a distinct participant. The partitioning was designed to reflect realistic scenarios where parties might hold complementary subsets of features for the same set of instances.  In this configuration, one party was identified as the active party (Guest), which possesses the labels essential for the learning task. The other parties were designated as passive (Hosts), each holding a slice of information about the same samples.\\\\ The experimental setup consisted of an active party $A$ and $n$ passive parties $[H_{1},.., H_{n}]$. Each local dataset was split into training and test sets, with 70\% of the data used for training and 30\% for testing. A vertical federated logistic regression model, as outlined in \cite{zhu2021federated}, was employed for training where no exchange of raw data occurs among parties, rather learning is achieved by exchanging intermediate results.  The F1-score, which harmonizes precision and recall, was chosen as the evaluation metric to assess the model performance. Furthermore, we experimented with a varying number of passive parties $n=3,4,5$ to briefly demonstrate the scalability of our method.
\subsection{Constructing Bankruptcy Game}
To construct the entire bankruptcy game, we began by calculating the estate \(E\) and the claims of the passive parties, \(c_{H_1}\)...\(c_{H_n}\), using the \cref{estate_eqn,claims_eqn}. These claims reflect the contributions of each passive party to the enhanced performance of the federated model compared to the local model of \(A\). According to the bankruptcy game, the value of any coalition is determined by the remaining estate after fulfilling the claims of parties not included in the coalition (\cref{bankruptcy_eqn}). We systematically compute these values for all possible coalitions of the passive parties, essential for analyzing the incentive distribution among participants. It is important to note that while our approach for calculating the nucleolus---ultimately the allocation vector in the bankruptcy game---does not inherently necessitate the computation of all possible coalition values, we perform this computation to facilitate a comparative analysis with the Shapley value of the same bankruptcy game. Unlike the nucleolus, the Shapley value computation requires considering all potential coalitions, making this exhaustive calculation informative for contrasting the two solution concepts.
\section{Results}
In this section, we analyze our experimental results (\cref{fig:results1}). We calculate the nucleolus (payoffs) of the bankruptcy game for the VFL incentive allocation task, compare estate claims and payoffs for passive parties, and also compare them with Shapley values (\cref{fig:results2}). The goal of this analysis is to systematically examine how alterations in the number of claimants and the magnitude of the estate influence the derived allocations. Each dataset was subjected to scenarios with 3 to 5 passive parties (hosts) to observe the distribution dynamics under different collaborative scales.
\begin{figure*}[t!]
  \centering
  \includegraphics[width=\textwidth]{plot_1_updated.png}
\caption[Overview of Estate, Claims, and Nucleolus Payoffs]{A comparison of the estate, claims, and nucleolus-derived payoffs for different host configurations across multiple datasets. The grey bars represent the total estate available in each scenario, the hatched portion within the estate denotes the claims of the hosts, and the orange bars illustrate their corresponding payoffs computed through the nucleolus. The sum of claims for each configuration is provided at the top of the subplots, labeled as $\Sigma C_j$.}
  \label{fig:results1}
\end{figure*}
\begin{figure*}[t!]
  \centering
  \includegraphics[width=\textwidth]{plot_2_updated.png}
   \vspace{-10pt}
\caption[Distribution of Payoffs Across Hosts in VFL Setting]{A detailed plot depicting the distribution of payoffs among hosts in a Vertical Federated Learning environment across different datasets. Each group of three bars represents a unique host configuration and their corresponding claims, nucleolus, and Shapley value allocations. The grey bars signify the claims of the hosts, while the orange and green bars illustrate the respective payoffs calculated using the nucleolus and Shapley value methods.}
\label{fig:results2}
\end{figure*}

Across all datasets, an increase in the number of hosts invariably leads to an increase in the sum of claims. This increment reflects the expanding demands on the estate as more claimants are involved. However, the nucleolus allocations maintain a proportional relationship with the claims, i.e, higher claims receive higher payoffs and vice versa,  ensuring that no party is significantly advantaged or disadvantaged, adhering to the principles of fair distribution within the constraints of the available estate. It is also consistently observed that, most claimants receive payoff at least half of their claim when the estate is sufficient. This outcome is in accordance with the principle of the Talmudic division rule, which stipulates equal awards to all claimants until each has received half of their claim. This rule ensures that when the estate can cover these minimum thresholds, claimants are guaranteed a fair share that reflects at least a portion of their stated claims. 

Annotations in the \cref{fig:results1,fig:results2} indicate the multiple by which the sum of claims exceeds the estate, such as \(\sum C_i = xE\), highlighting the pressure on the estate with an increasing number of claims. Specific observations in individual datasets reveal interesting patterns such as the Synthetic and Bank Marketing datasets illustrate scenarios where the aggregate claims substantially exceed the estate's value, especially with 5 hosts. A key observation is that, When total claims significantly surpasses the estate, payoffs reduce, revealing the limitation of the estate to satisfy all claims and necessitating equitable payout adjustments. Particularly in scenarios like the Bank Marketing dataset with 5 hosts, once claims surpass a certain threshold (Hosts H3, H4, H5), payouts become uniform, adhering to the principles of Talmudic division. On the other hand, when the total claims are equal or less than the estate, all claims are met, occasionally leaving a surplus. For instance, in the 3-host configuration for the Heart Disease dataset, H1's claim is fully satisfied with excess estate remaining, whereas H2 and H3, having zero claims, rightly receive no payoff. Similarly, for 5 hosts, H3 and H4 with zero claims receive no part of the estate. This demonstrates the rule that claimants with no contribution receive nothing.

\cref{fig:results2} demonstrates a comparison between the computed Shapley values and nucleolus allocations for the same bankruptcy game in our VFL scenario. Despite the differences in values—attributable to the distinct fairness concepts inherent to the nucleolus and Shapley methods—the allocations are found to be comparable. This comparison highlights that although the Shapley values and nucleolus allocations are not exactly equal, their relative distributions across claimants bear similarities, reinforcing the feasibility of using the nucleolus as a computationally efficient alternative to the Shapley value for fair incentive allocation in a bankruptcy game. 
\subsection{Interpreting Payoffs in Practical Setting}
In our proposed approach, the payoffs obtained by each party can be compared with the estate to compute what portion of the estate it should receive. This percentage can then be utilized to allocate incentives in practical settings, especially when there's a predefined budget for the federation. Taking the experimental results for the spam email dataset when \(n=3\) as an instance, the percentages of the payoffs for \(H1\), \(H2\), and \(H3\) would be approximately \(43.5\%\), \(17.7\%\), and \(38.8\%\) respectively, computed as \(\frac{64.0}{147.0} \times 100\), \(\frac{26.0}{147.0} \times 100\), and \(\frac{57.0}{147.0} \times 100\). Given these calculations, if the total budget set aside by the active party for this federation amounts to, let's say, \$10,000, then \(H1\) would receive \$4,350 (43.5\% of the budget), \(H2\) would secure \$1,770 (17.7\% of the budget), and \(H3\) would be allocated \$3,880 (38.8\% of the budget).
 Hence, this method can ensure that the participating parties are incentivized to contribute high-quality and unique data while also ensuring that the incentive budget is allocated fairly among the parties. 

\subsection{Fairness Evaluation}\
In game theory, certain principles ensure equitable allocation of benefits among players, serving as benchmarks for justifiable distribution in cooperative games denoted as \emph{``fairness axioms''} (cf. \cite{roth1988introduction}). Both nucleolus and Shapley value adhere to the fairness axioms: Efficiency, Null Player, and Symmetry \cite{peters2015game}. In this section, we have explored the concept of fairness in our incentive allocation approach using Nucleolus as well as for Shapley values by examining the fairness axioms. To empirically evaluate these principles, we utilize the Spam Email dataset, selecting a scenario that involves one active party and three passive parties, designated as \textit{H1}, \textit{H2}, and \textit{H3}. 

\begin{table*}[ht]
\centering
\footnotesize
\setlength{\tabcolsep}{0.06cm}
\begin{tabular}{l|l|llll|llll|llll}
\hline
\multicolumn{2}{c|}{} & \multicolumn{4}{c|}{Claims} & \multicolumn{4}{c|}{Nucleolus} & \multicolumn{4}{c}{Shapley Value} \\
\hline
Property & Estate & H1 & H2 & H3 & H4 & H1 & H2 & H3 & H4 & H1 & H2 & H3 & H4 \\
\hline
Efficiency & 147 & 106 & 52 & 99 & - & 64.0 & 26.0 & 57.0 & - & 59.17 & 32.17 & 55.67 & - \\
Null Player & 144 & 106 & 52 & 99 & 0 & 62.5 & 26.0 & 55.5 & 0.0 & 58.17 & 31.17 & 54.67 & 0.0 \\
Symmetry & 147 & 106 & 52 & 99 & 99 & 40.33 & 26.0 & 40.33 & 40.33 & 42.42 & 24.42 & 40.08 & 40.08 \\
\hline
\end{tabular}
\caption{Empirical Evaluation of Fairness Axioms, illustrating the claims, and allocations from both Nucleolus and Shapley Value methods, for properties of Efficiency, Null Player, and Symmetry.}
\label{tab:fairness_axioms}
\end{table*}

\noindent \textbf{Efficiency: } The total payoff distributed among players is equal to the game's total value or grand coalition. Mathematically given by $\sum_{i \in N} \phi_i(v) = v(N)$, where $N$ represents the set of all players, $\phi_i(v)$ denotes the payoff allocated to player $i$, and $v(N)$ is the value of the grand coalition. 
\noindent From the \cref{tab:fairness_axioms}, we observe that, both the Nucleolus and Shapley Value methods allocate improvements from a federated learning scenario, where the estate, valued at 147 units, signifies the overall improvement in model performance through VFL. This estate is distributed among three passive parties (\textit{H1, H2, H3}). The Nucleolus allocations are 64.0, 26.0, and 57.0 units, while the Shapley Value assigns 59.17, 32.17, and 55.67 units to \textit{H1}, \textit{H2}, and \textit{H3}, respectively. Both methods adhere to the efficiency axiom, ensuring all available improvement is allocated precisely, with no surplus.

\noindent \textbf{Null player: } If a player doesn't change the value of any coalition by joining or leaving, i.e. \( v(S \cup \{i\}) = v(S) \quad \forall S \), then \( \phi(i) = 0 \). In Talmudic division rule, by definition, claimants having zero claims (\(C_i = 0\)) are disregarded, and no allocation is made to them (\(x(i) = 0\)). 

\noindent To demonstrate the null player property, we introduce an additional passive party, \textit{H4}, into the federation. The data attributed to \textit{H4} was randomly generated and deliberately designed to bear no relevance to the learning task at hand, thus theoretically contributing no value to the federated model or the grand coalition. When constructing the bankruptcy game for incentive allocation in VFL, claim of \textit{H4} is, by definition, zero, as its participation does not contribute to any improvement in the model performance alongside the guest party.  According to Talmudic division, claims with zero value are inherently disregarded and no part of estate is allocated to it which aligns with the null player property of cooperative game theory. This property could be crucial in addressing one of the challenges in VFL - identifying malicious hosts. Moreover, by leveraging the null player property, we can effectively isolate and mitigate the impact of parties that contribute irrelevant or detrimental data to the federated learning process.

\noindent \textbf{Symmetry: } If two players, \(i\) and \(j\), contribute identically to any coalition \(S \subseteq N \setminus \{i, j\}\) they are part of, their allocated payoffs should be equal, formally expressed as \[ \phi_i(v) = \phi_j(v) \] This property is equivalent to the Talmudic division rule property, Equal Treatment of Equals (ETE) \cite{thomson2003axiomatic}. The Talmudic rule explicitly states that two claimants with identical claims should receive identical allocations. This is succinctly captured by the condition and consequence relationship \[ c_i = c_j \implies x_i = x_j \] Here, \(c_i\) and \(c_j\) denote the claims of players \(i\) and \(j\), respectively, while \(x_i\) and \(x_j\) represent their respective allocations. 

\noindent For examining the symmetry property, we introduce a fourth passive party, \textit{H4}, into the federation, similar to the previous scenario. However, in this case, the data attributed to \textit{H4} is identical to that of \textit{H3}. According to our approach, the claims of \textit{H3} and \textit{H4} should be computed to be the same due to their identical contributions to the learning task. Consequently, the allocations for both \textit{H3} and \textit{H4}, determined by both the Nucleolus and Shapley methods, will be equal. This outcome satisfies the symmetry property (\cref{tab:fairness_axioms}), reflecting their symmetrical impact on model improvement within the VFL setting. However, it is important to note that while the symmetry property can sometimes be useful for addressing redundant data issues in a VFL setting, it does not always guarantee redundancy detection. This is because the property cannot distinguish whether the data from two parties is redundant or if both hosts have different data of equal worth. This goes for the Shapley values as well. Therefore, while the symmetry property can help treat similar or identical data fairly, additional mechanisms for participant selection \cite{jiang2022vf} are necessary to detect and manage redundant data in VFL accurately.

\subsection{Computational Complexity}
\begin{figure}[t!]
  \centering
  \includegraphics[scale=0.28]{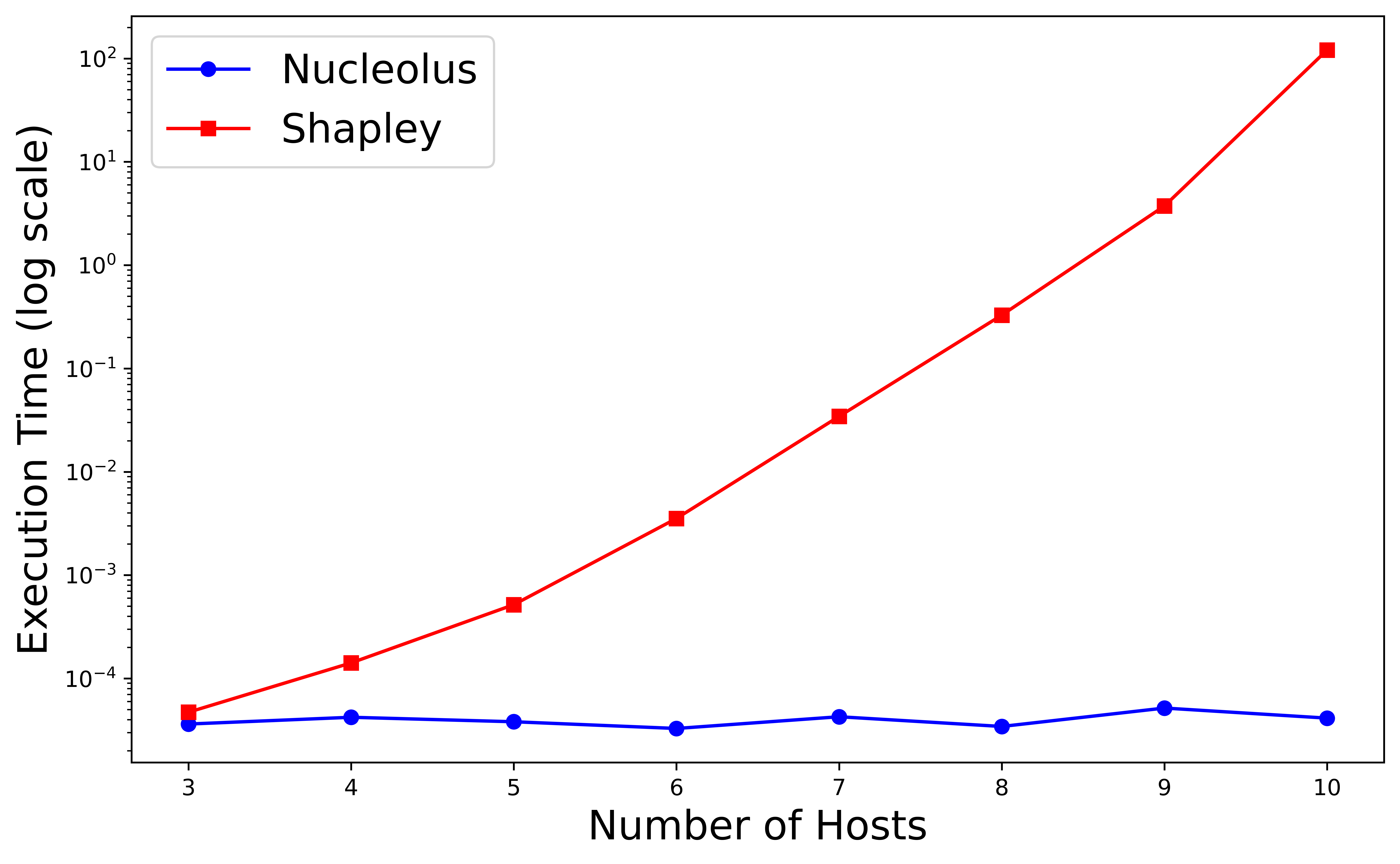}
  \caption{Execution Time Comparison Using Nucleolus vs Shapley Value on Spam Email Dataset with $10$ Passive Parties}
  \label{fig:results3}
\end{figure}

In bankruptcy games, the Shapley value requires computation across all possible coalitions, facing a combinatorial explosion as the number of participants increases, with its computational complexity being equal to $2^n$. This makes it less feasible for large-scale applications within bankruptcy scenarios. On the other hand, the nucleolus sidesteps the exhaustive coalition computations when it is computed using Talmud division rule, achieving equivalent allocations without necessitating the computation of all coalition values, thus operating within a linear complexity of $O(n)$ (\cref{fig:results3}). This significant computational difference makes nucleolus suitable for scenarios with a large number of participants in a vertically federated setting. Our comparative analysis (\cref{fig:results2}) demonstrates minor differences between the Shapley value and Nucleolus payoffs. Hence nucleolus ensures both computational efficiency and adherence to fairness principles, aligning with the needs of large-scale VFL environments.

\section{Conclusion and Future Direction}

In this paper, we tackle the challenge of achieving fair and efficient incentive allocation among passive parties in Vertical Federated Learning by conceptualizing it as a bankruptcy game. Our proposed methodology, based on the solution concept \emph{nucleolus} computed through the Talmudic division rule, ensures equitable incentive distribution. Moreover, our approach preserves privacy since raw data never leaves its original location, addressing a fundamental concern in federated learning. Through extensive experimentation across various datasets, we have proved that our approach can adapt to diverse scenarios without imposing significant computational burdens. This renders it particularly suitable for large-scale federated learning contexts with multiple data parties. However, a limitation of our method is that it considers only the coalitions formed between the active party and individual passive parties. As a result, non-linear dynamics among passive parties do not emerge. While this simplification reduces computational complexity and mitigates privacy risks associated with evaluating every possible coalition, it also aligns with the practical reality that forming artificial coalitions among passive parties may not be necessary or meaningful in most federated learning applications.

VFL is particularly useful for collaboration among non-competing parties, where data privacy is important. This type of collaboration is useful in industries such as finance and marketing. For example, in finance, VFL enables banks or financial institutions to collaborate on improving fraud detection systems while maintaining the privacy of customer data. In marketing, companies can work together on consumer behavior analysis without exposing sensitive information, allowing for the development of more effective marketing strategies. For these federations to run smoothly, establishing proper incentives for participation is essential. Our proposed method for incentive allocation can be applied effectively in these cases due to its simple yet fair nature, ensuring that all parties are motivated to contribute high-quality data while preserving privacy. Looking ahead, future research could explore the integration of a party selection mechanism within VFL prior to the training and incentive allocation processes by privacy-preserving data quality analysis, potentially enhancing the overall efficiency of the incentive distribution framework. Although we tested our method on real-world data sets, further research would involve exploring this method on more diverse datasets or through an application in practice.


\bibliography{ecai}

\end{document}